\documentclass[letterpaper, 10 pt, conference]{ieeeconf}  

\IEEEoverridecommandlockouts                              
\usepackage{graphics} 
\usepackage{epsfig} 
\usepackage{mathptmx} 
\usepackage{times} 
\usepackage{amsmath} 
\usepackage{amssymb}  

\usepackage{subfig}
\usepackage{caption}

\title{\LARGE \bf Team Applied Robotics: A closer look at our robotic picking system}

\author{Wim Abbeloos$^{1}$, Fabian Gouwens, Simon Jansen, Berend K\"{u}pers, Maurice Ramaker, Toon Goedem\'{e}$^{1}$%
\thanks{$^{1}$Wim Abbeloos and Toon Goedem\'{e} are with KU Leuven, Department of Electrical Engineering, Leuven, Belgium}
\thanks{The authors thank Alten, KU Leuven and Smart Robotics for their support.}
}
\begin{document}
\maketitle
\thispagestyle{empty}
\pagestyle{empty}

\begin{abstract}

This paper describes the vision based robotic picking system that was developed by our team, Team Applied Robotics, for the Amazon Picking Challenge 2016.  This competition challenged teams to develop a robotic system that is able to pick a large variety of products from a shelve or a tote.  We discuss the design considerations and our strategy, the high resolution 3D vision system, the use of a combination of texture and  shape-based object detection algorithms, the robot path planning and object manipulators that were developed.

\end{abstract}

\section{INTRODUCTION}
There is a strong interest in applying robotics not only in the typical, well structured environments they normally operate in, but also in less structured, real world situations.  While there has been a lot of progress in the fields of computer vision, path planning and robotic grasping, combining them into a reliably working system still proves to be challenging.  The APC is a competition in the field of warehouse logistics, in which objects need to be picked either from a shelve, or a tote.

\begin{figure}
\centering
\includegraphics[width=0.35\textwidth]{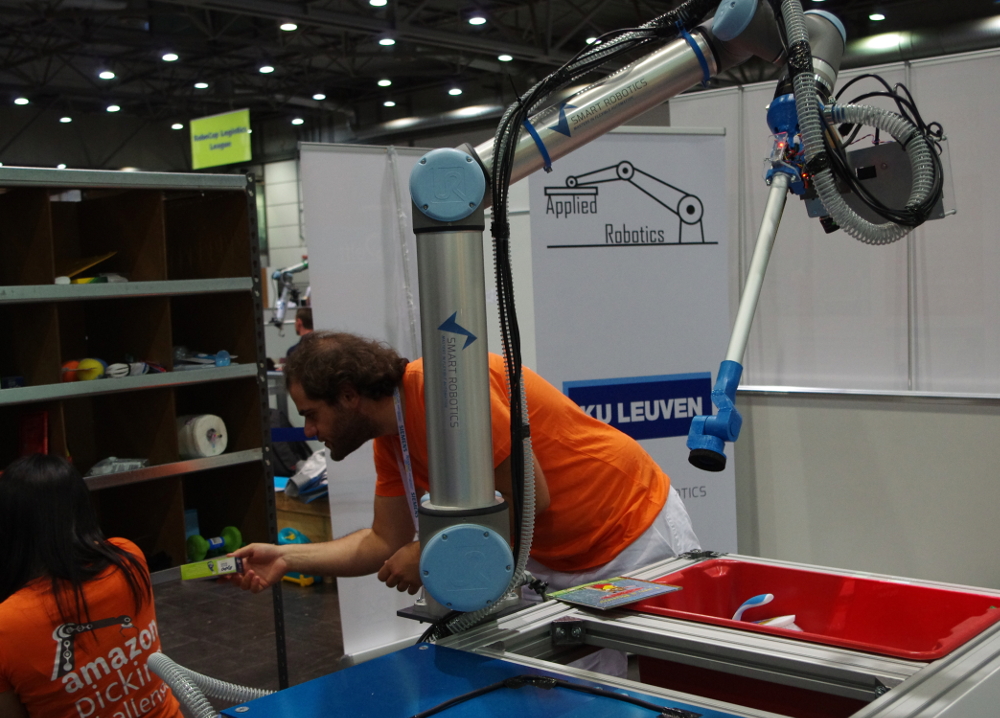}
\caption{The developed picking system consists of a standard six degree of freedom robot equipped with a custom made 3D vision system and vacuum gripper.}
\label{fig:robot}
\end{figure}

The system we built is based on a Universal Robots UR10 robot (Figure~\ref{fig:robot}).  The system was outfitted with a custom vision sensor and vacuum gripper.  The software ran on a single laptop and made use of the Robot Operating System (ROS) framework.  In the following sections we provide more details on the vision system, object detection, path planning and object manipulation.

\section{Perception}

A 3D vision system was developed to detect and determine the 6D pose of the objects that need to be handled.

\subsection{Time Multiplexed Structured Light}

\begin{figure}
\centering
\includegraphics[width=0.35\textwidth]{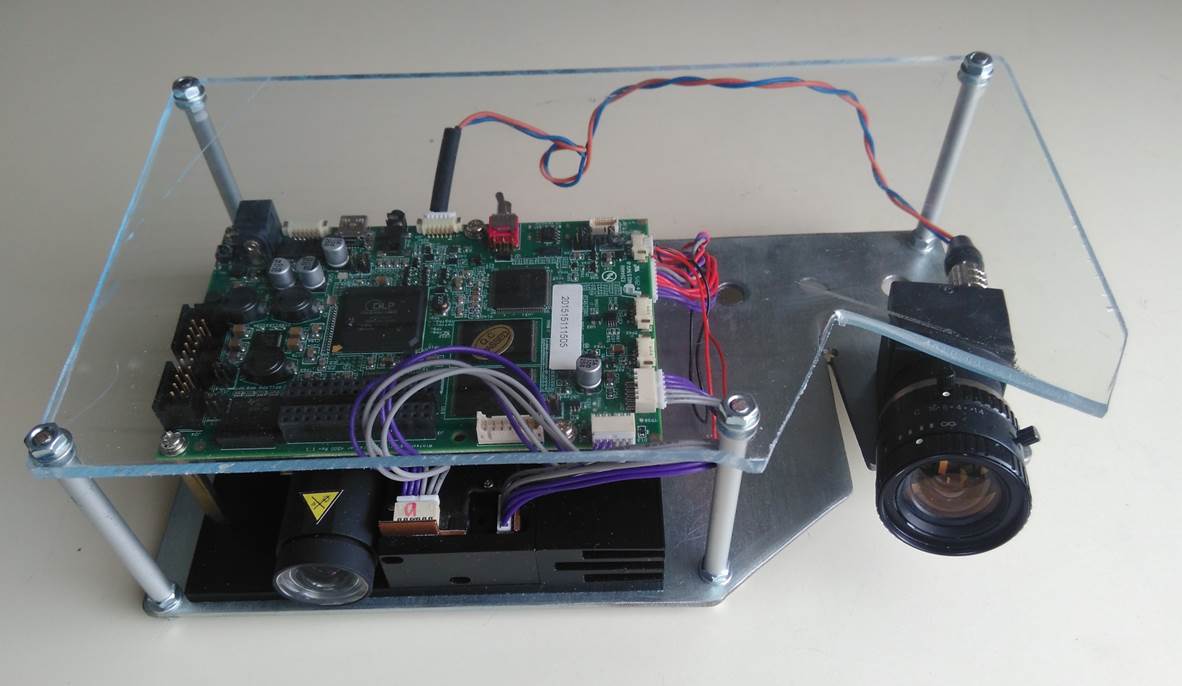}
\caption{The 3D vision system consists of a Digital Micromirror Device projector (TI DLP lightcrafter 4500) and a monochrome camera (Pointgrey Flea3 FL3-U3-13Y3) equipped with a c-mount lens (Computar 8mm F1.4 M0814-MP2).  The mount was made from a bent steel plate and attaches directly to the robot wrist.  A protective acrylic top shields the components.  The projector and camera are synchronized to ensure the camera captures the correct projected patterns.}
\label{fig:sensor}
\end{figure}

To be able to recognize objects and determine their pose with sufficient accuracy, while keeping system cost low, a custom 3D scanner (Figure~\ref{fig:sensor}) was built from off the shelve components.   The system is based on triangulation between binary Gray code  pattern sequences projected by a projector, and their image as acquired by a camera~\cite{inokuchi1984range}\cite{posdamer1982surface}.  The camera and projector are synchronized and the patterns can be projected and acquired at 120 frames per second.

Our choice for this custom sensor over a standard depth sensor such as Time of Flight cameras, the Microsoft Kinect, or the Intel RealSense was mostly based on the higher resolution (1140x912 pixels) and accuracy (0.1mm).  The biggest downside is that it requires multiple images (42), and some extra processing.  This results in a longer acquisition time (about one second for acquisition and processing).

\subsection{Sensor Calibration}
To be able to triangulate points, both the camera and the projector need to be calibrated.  The camera is calibrated in the typical way, using a chessboard pattern.  Note that while the projector can be modeled as a camera, it is not possible to measure the location of the checkerboard corners in the projector reference frame directly.  The method proposed by Moreno et al. \cite{moreno2012simple} was used to estimate the position of the corners in the projected image.  This method uses the decoded pattern as observed by the camera and creates a local homography to estimate the code that can be associated to the checkerboard corner.  This allows the projector to be calibrated, and to calibrate the camera-projector setup as a stereo camera.

\section{Detection}

A set of 38 products of varying shape, appearance, material and weight was used in the challenge.  To cope with this variety, we chose to employ multiple detection algorithms, and apply the most suitable for every object.  One object was searched for at a time, starting with the object that was deemed easiest, according to an ordered list that was created manually (taking into account ease of detection and ease of manipulation).  Also, for every object, a preferred object detection method was determined in advance.

For picking from the shelve, three scans were taken from different angles.  For picking from the tote, two scans were taken from different angles.  After picking it from the tote, the item was placed on a table and an additional scan was taken in order to confirm it was the correct object.

\subsection{Pre-processing}

\begin{figure}
   \centering
   \subfloat[Intensity image]{{\includegraphics[width=4cm]{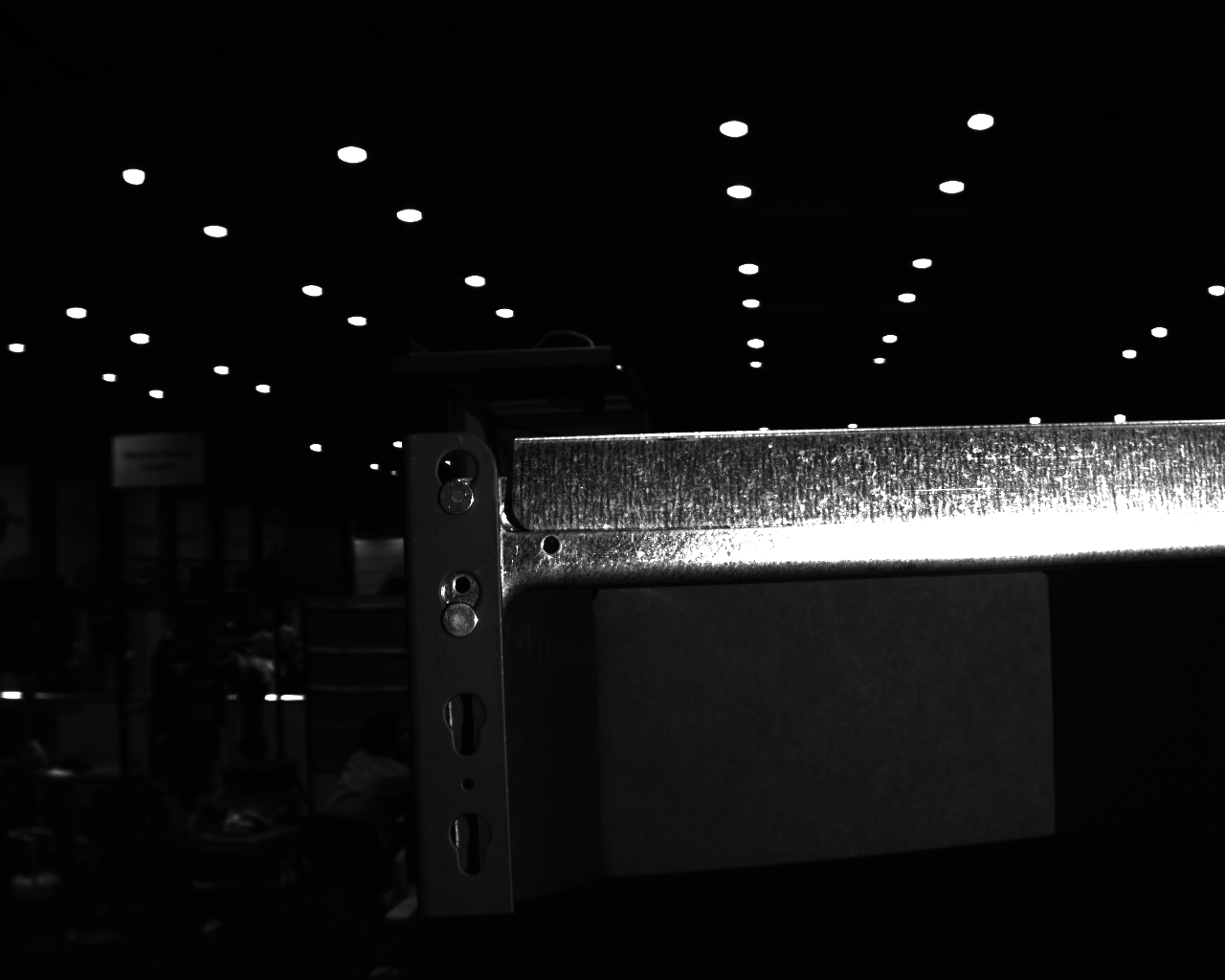} }}%
    \subfloat[Range image]{{\includegraphics[width=4cm]{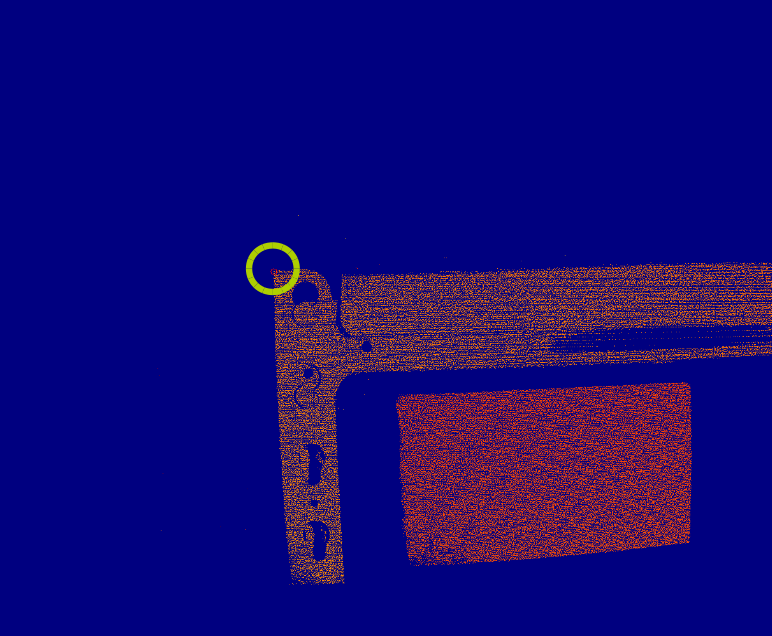} }}%
    \caption{A scan of the top corners of the shelve is used to measure the pose of the shelve with respect to the robot.  The detected corner is shown in the range image with a green circle.}%
    \label{fig:corner}%
\end{figure}

As the shelve and tote geometry were known, it is possible to segment the objects from the background.  While the exact position of the tote was known, the shelve was slightly moved before the start of the competition.  A calibration routine was used at the start of the competition in which the shelves top corners were scanned (Figure~\ref{fig:corner}) and their exact position was measured, allowing to determine the shelve pose.

\subsection{Detection Algorithms}
\subsubsection{Point Pair Features}
\begin{figure}
\centering
\includegraphics[width=0.25\textwidth]{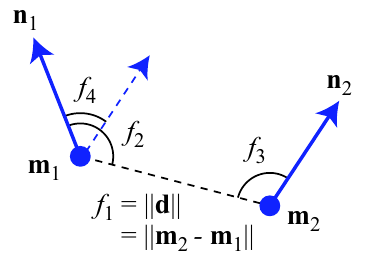}
\caption{The four dimensional point pair feature vector $\bf{f}$ of two 3D points on the surface of an object: $m_{i}$, and their normals $n_{i}$.  The first element, $f_{1}$ is a distance, while $f_{2}$, $f_{3}$ and $f_{4}$  are angles.}
\label{fig:ppf}
\end{figure}

Some of the objects have a distinctive geometric shape.  Point pair features (PPF, Figure~\ref{fig:ppf}) can be used to describe this shape \cite{Abbeloos16ppf}\cite{choi2012voting}\cite{wahl2003surflet}.  While a single feature is not very descriptive, the features of all combinations of points on the objects surface typically is.  

From a measured scene, a subset of points is selected, and the PPFs of all their combinations are calculated.  If a similar PPF is present in the objects model, the PPF votes for a 6D pose of the object.  If a pose gets enough votes, it is accepted as a detection.  The initial detection is followed by an Iterative Closest Point (ICP) procedure to obtain a refined object pose.

\subsubsection{2D Features}

\begin{figure*}
   \centering
   \subfloat[Grayscale image]{{\includegraphics[width=5.5cm]{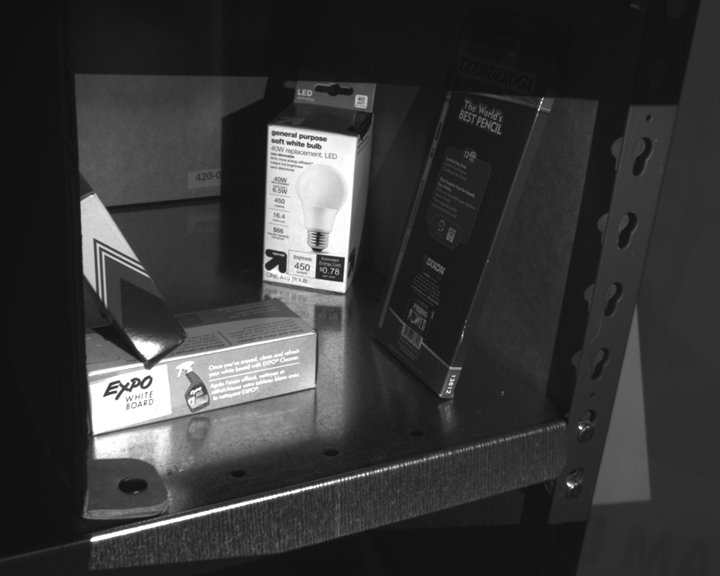} }}%
    \quad
  \subfloat[Filtered pointcloud]{{\includegraphics[width=5.5cm]{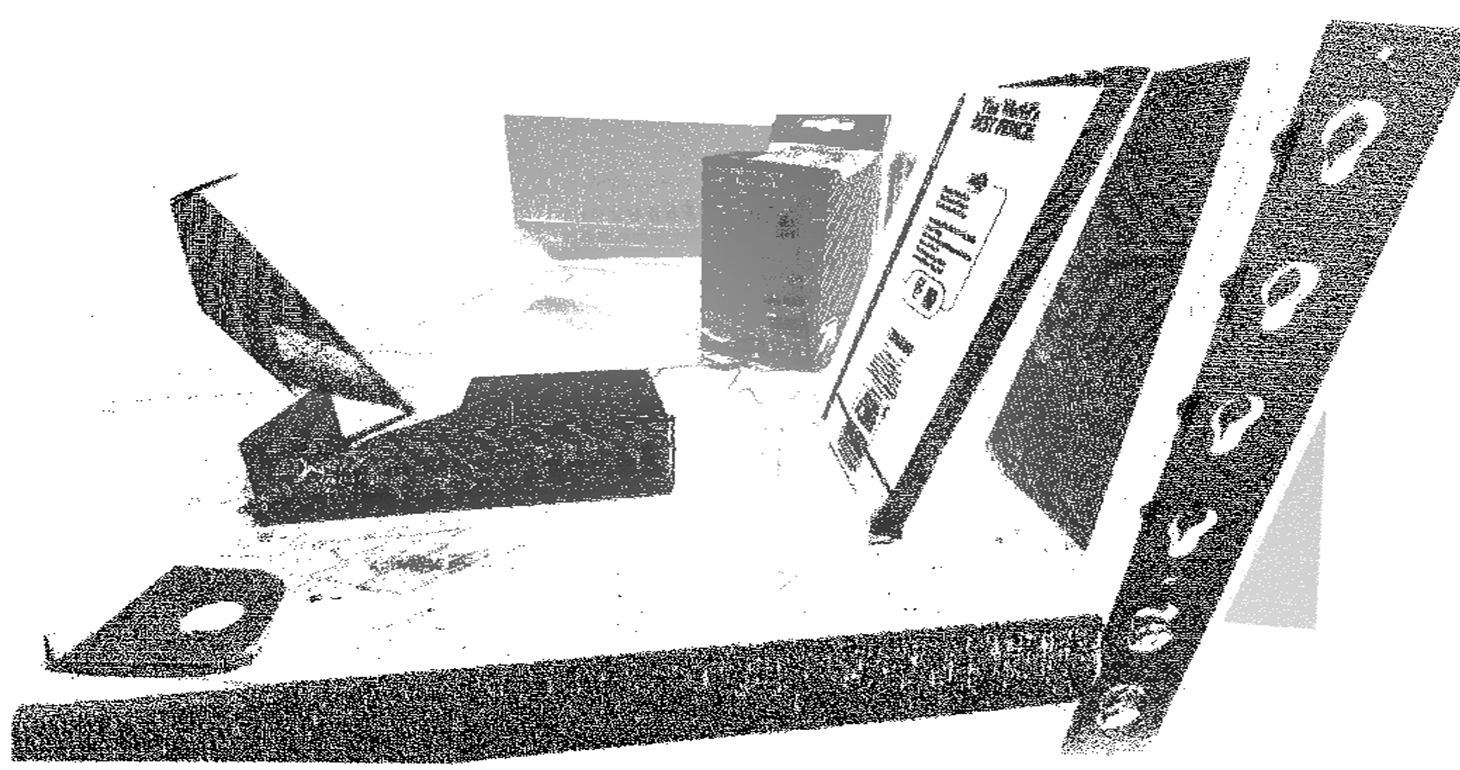} }}%
    \quad
    \subfloat[Matched object]{{\includegraphics[width=5.5cm]{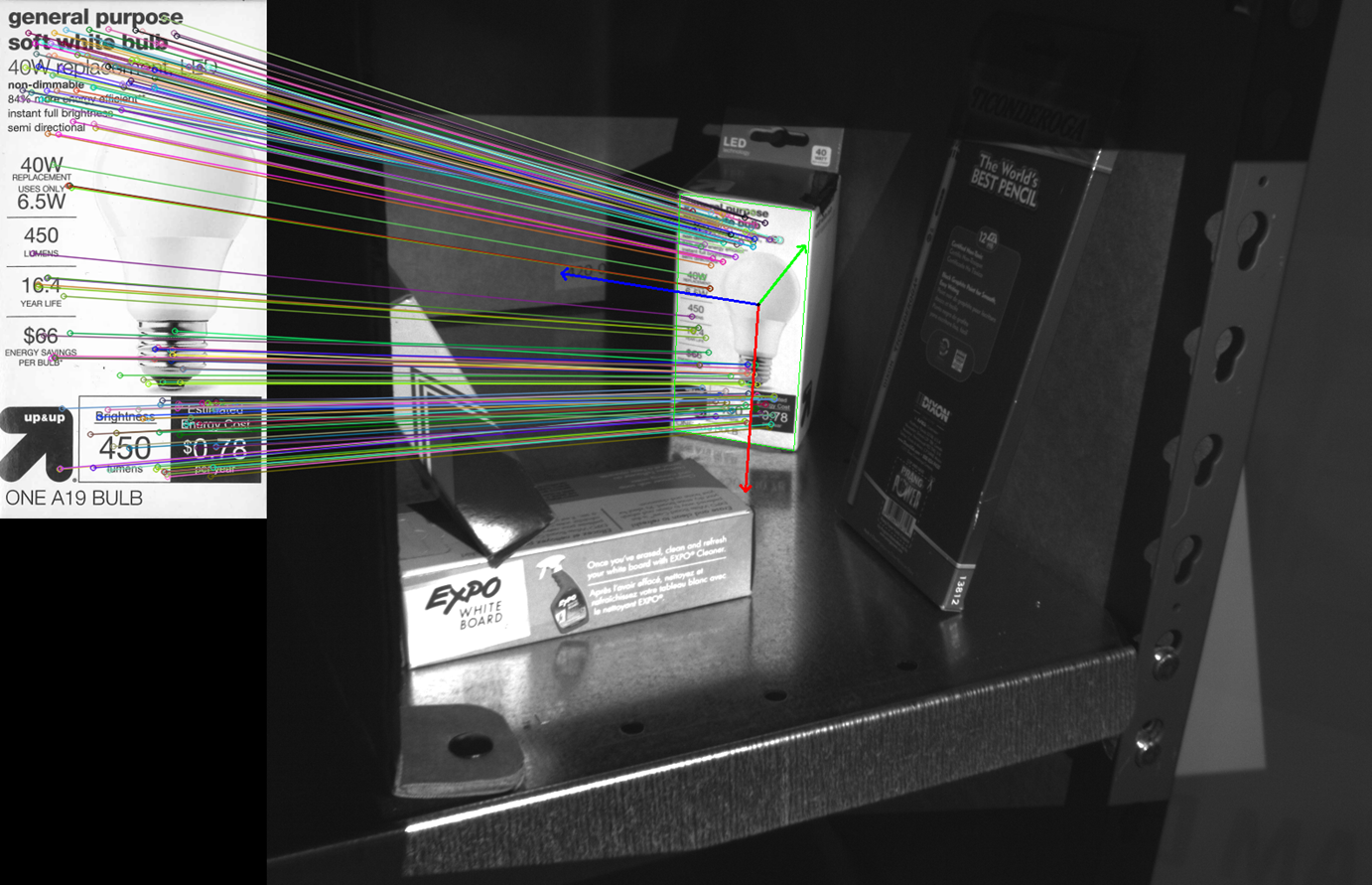} }}%
    \caption{Local appearance features of the objects are matched to the scene, allowing to estimate the objects pose.  Using the 3D location of the feature point, from the pointcloud, results in higher accuracy and allows to eliminate incorrect matches.}%
    \label{fig:feature}%
\end{figure*}

The objects with distinctive texture can be detected using a local features based approach \cite{collet2009object}\cite{grundmann2010robust}, in our case, SIFT is used as a feature descriptor.  From these matches between the object template and scene (Figure~\ref{fig:feature}) the pose can be estimated using a Perspective-n-Point algorithm, in this case EPnP~\cite{lepetit2009epnp} was used.  Note that in our case, the exact 3D location of all features is measured with the structured light scanner.  This additional information allows to estimate the pose with much higher accuracy, and allows to eliminate incorrect matches.

\subsubsection{Range Image Templates}
\begin{figure}
   \centering
   \subfloat[Scene]{{\includegraphics[width=5cm]{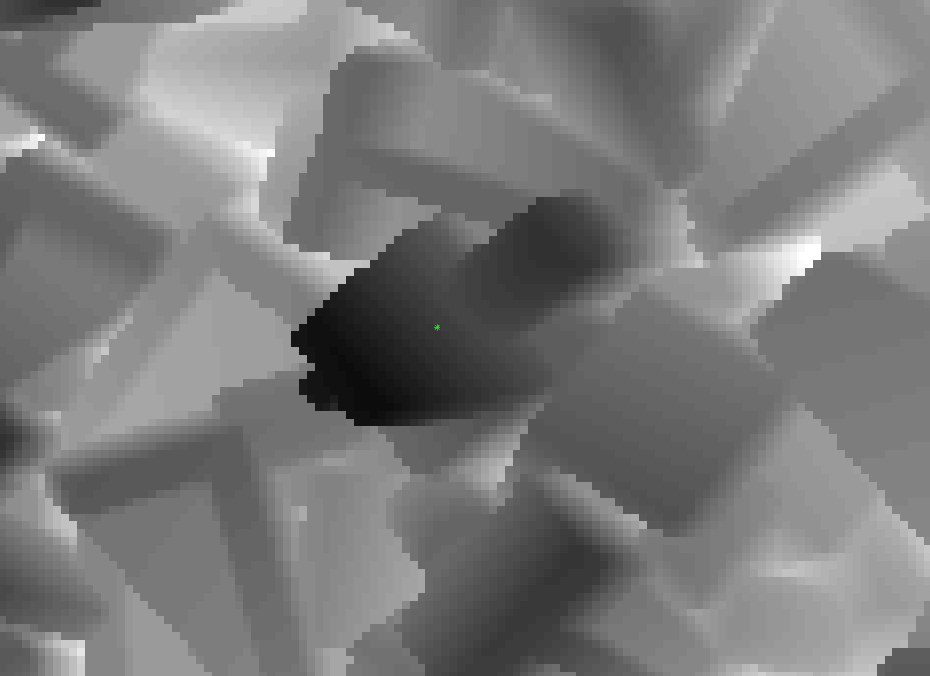} }}%
    \subfloat[Selected template]{{\includegraphics[width=3cm]{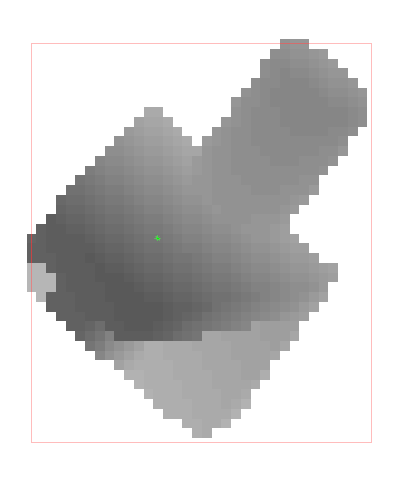} }}%
    \caption{A set of 1024 templates (range images from different viewpoints) are aligned to the scene using an optimization function.  The final pose of the template with the lowest cost is used if it is below a certain threshold.}%
    \label{fig:Range}%
\end{figure}
The measured pointcloud can also be represented as a range image (Figure~\ref{fig:Range}).  If the above algorithms fail, a brute force approach can be used in which templates obtained from the model are compared to this range image \cite{germann2007automatic}.  We use 1024 templates per object.  Each of the templates is aligned to the scene in an optimization procedure.  To initialize the optimization, the scene is first segmented using euclidean clustering, and the closest point of each of the clusters can be used as a starting point, which is initially aligned to the closest point in the range template to be optimized.  If a template with sufficiently low cost is found, an extra ICP step is used to refine the object pose.

\section{Manipulation}

\begin{figure}
\centering
\includegraphics[width=0.4\textwidth]{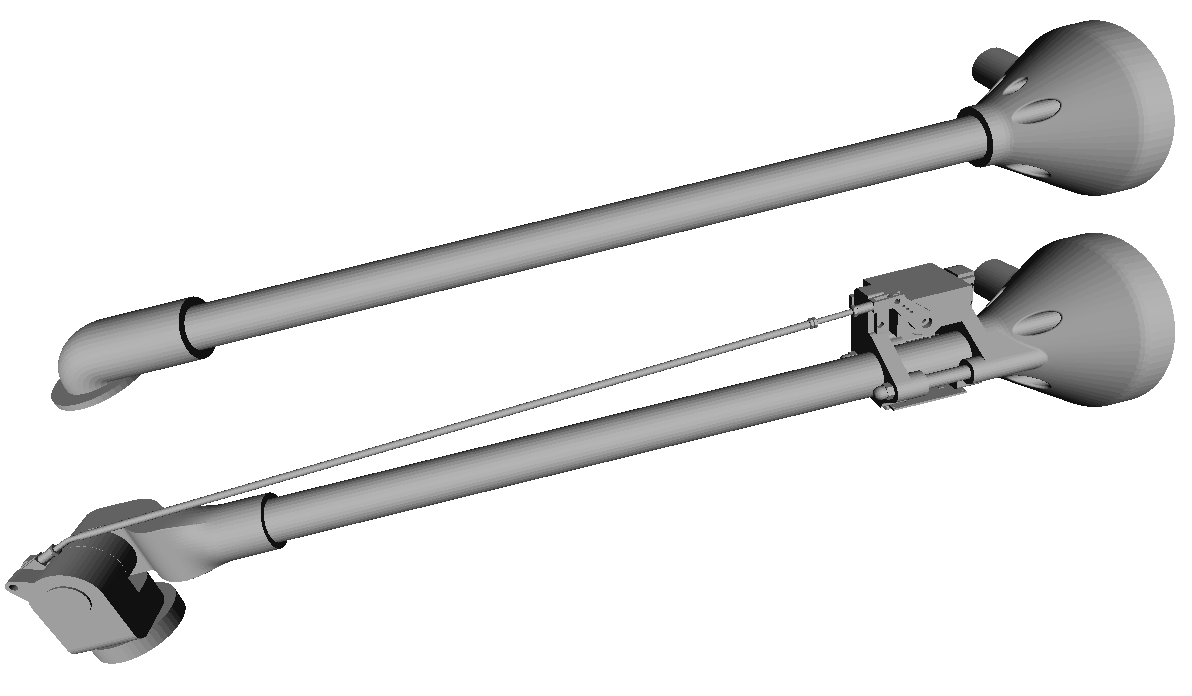}
\caption{\textbf{Top:} The gripper used for picking from the shelve. \textbf{Bottom:} The gripper used for picking from the tote has an extra degree of freedom.  The 3D printed joint is designed to have a bit of play, but seals itself when the vacuum is turned on, so that it does not leak air.  The joint can be actuated by a servo motor via a bar linkage.  The servo motor is controlled by a micro controller with an accelerometer and has three modes: unactuated, fixed at a certain angle, or set to keep the head vertical.}
\label{fig:gripper}
\end{figure}

Two suction based grippers were custom designed (Figure~\ref{fig:gripper}): one for picking from the shelve and one for picking from the tote.  They both consist of pieces of standard aluminum tube and parts made using an additive manufacturing process (fused deposition modeling).  

Both grippers use high flow vacuum generated by a modified 2000W vacuum cleaner.  The air flow is guided through flexible tubing.  A small circular piece of soft foam was glued to the grippers to provide better sealing of the gripper to the object.

When grasping an object, the end effector was lowered until a certain force threshold was met.  This increased the chance of grasping the object properly.

\begin{figure*}
   \centering
   \subfloat[Pointcloud and detected object]{{\includegraphics[width=5.5cm]{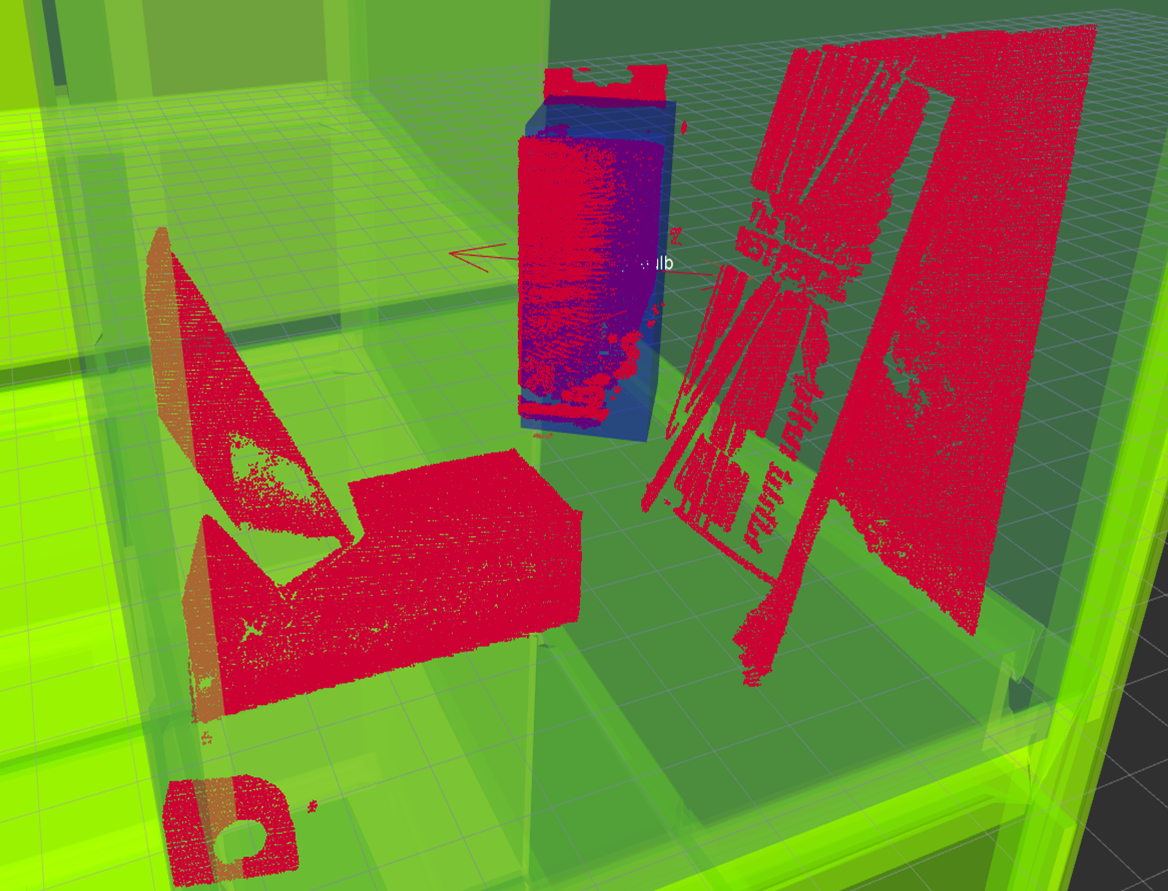} }}%
    \quad
  \subfloat[Octomap approximation]{{\includegraphics[width=5.5cm]{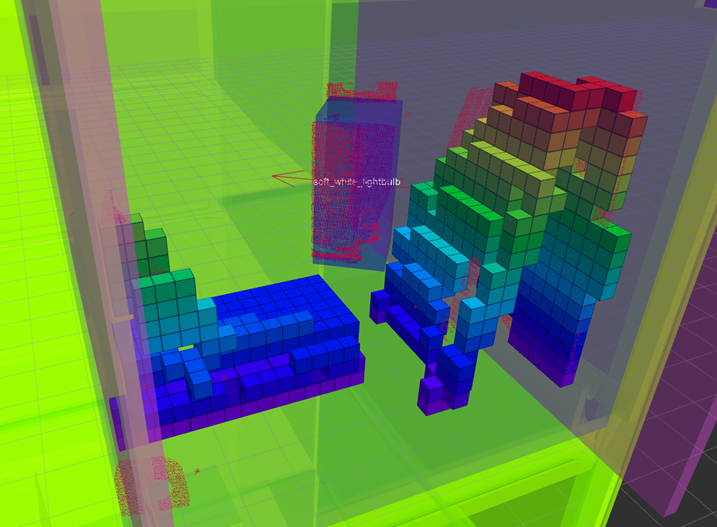} }}%
    \quad
    \subfloat[Planned path]{{\includegraphics[width=5.5cm]{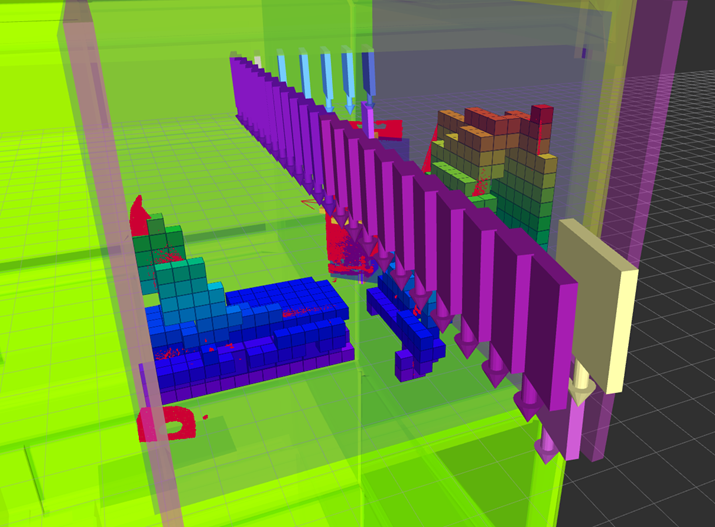} }}%
    \caption{When an object is detected, a collision map is calculated using on octomap approximation of the pointcloud.  Note that the points on the objects are not included.  These point are cut out of the pointcloud using the a slightly expanded version of the convex hull of the object.  A cartesian path planning algorithm is used to search for a collision free path to remove the object from the shelve. }%
    \label{fig:planning}%
\end{figure*}

\section{Planning}
Once an object is detected the object needs towards towards it to grasp it.  The grasping locations were manually defined per object.  Most of the robot movements were planned in advance, offline, in joint space.  Only the final part of the paths toward the object and out of the shelve need to be planned online.  This was done using a cartesian path planning, which creates a tree structure of cartesian movements, with a resolution of 2cm. A greedy search algorithm was used and a check for collisions performed at every node (Figure~\ref{fig:planning}).  For removing objects from the tote, no path planning was used and the object was simply lifted vertically.

\section{Conclusions}
Our approach to the challenges met in the APC was summarized in this paper.  The proposed system has some unique features that were not employed by any other teams.  Our system also suffered from some drawbacks, one of the major limitations during the challenge was that the 3D scan acquisition speed which had to be lowered due to technical issues.  Another limitation was that not all of the objects could be detected reliably.  The vacuum detection  did not work properly, and could not be used to tell whether an object was successfully picked.  Despite these issues, our system was able to pick multiple items correctly in both of the challenges.


\bibliographystyle{IEEEtran}
\bibliography{IEEEabrv,APC}

\end{document}